\def\year{2022}\relax
\documentclass[letterpaper]{article} 
\usepackage{aaai22}  
\usepackage{times}  
\usepackage{helvet}  
\usepackage{courier}  
\usepackage[hyphens]{url}  
\usepackage{graphicx} 
\usepackage{natbib}  
\usepackage{caption} 
\DeclareCaptionStyle{ruled}{labelfont=normalfont,labelsep=colon,strut=off} 
\usepackage{amsmath}
\usepackage{amssymb}
\usepackage{amsthm}
\usepackage{booktabs}
\usepackage{algorithm}
\usepackage{algorithmic}
\usepackage{subfig}
\usepackage[table,xcdraw]{xcolor}

\title{Hierarchical Representation Learning for Markov Decision Processes
}
\author{
    Lorenzo Steccanella \textsuperscript{\rm 1} \footnote{Contact Author}, 
    Simone Totaro \textsuperscript{\rm 2} , 
    Anders Jonsson \textsuperscript{\rm 1} \\
}
\affiliations{
    \textsuperscript{\rm 1}Dept.~Information and Communication Technologies, Universitat Pompeu Fabra, Barcelona, Spain\\
    \textsuperscript{\rm 2} Mila -- Quebec Artificial Intelligence Institute, Montreal, Canada \\
    
    \{lorenzo.steccanella, anders.jonsson\}@upf.edu, simone.totaro@mila.quebec
}

\begin{document}
\maketitle
\vspace{30pt}
\begin{abstract}
In this paper we present a novel method for learning hierarchical representations of Markov decision processes. Our method works by partitioning the state space into subsets, and defines subtasks for performing transitions between the partitions. At the high level, we use model-based planning to decide which subtask to pursue next from a given partition. We formulate the problem of partitioning the state space as an optimization problem that can be solved using gradient descent given a set of sampled trajectories, making our method suitable for high-dimensional problems with large state spaces. We empirically validate the method, by showing that it can successfully learn useful hierarchical representations in domains with high-dimensional states. Once learned, the hierarchical representation can be used to solve different tasks in the given domain, thus generalizing knowledge across tasks.
\end{abstract}

\section{Introduction}

In reinforcement learning, an agent attempts to learn behaviors through interaction with an unknown environment. By observing the outcome of actions, the agent has to learn from experience which action to select in each situation in order to maximize the expected cumulative reward. To learn, the agent has to balance exploration, i.e.~discovering the effect of actions on the environment, and exploitation, i.e.~repeating action choices that have proven successful in the past.

In hierarchical reinforcement learning~\cite{10.1023/A:1022140919877}, the task is decomposed into subtasks, and the solutions to the subtasks are combined to form a solution to the overall task. If each subtask is easier to solve than the overall task, the decomposition can significantly speed up learning. The subtasks can also help explore the environment more efficiently, since one high-level decision typically brings the learning agent multiple steps in a promising direction, rather than exploring locally one step at a time.

In most applications of hierarchical reinforcement learning, the subtask decomposition is provided by a domain expert that exploits extensive domain knowledge to define appropriate subtasks. Though many automatic methods have been proposed, learning a useful subtask decomposition from experience is still mostly an open research question. In addition, several of the proposed methods are not appropriate for high-dimensional problems since they maintain statistics about individual states.

In this paper, we propose a novel method for learning a hierarchical representation for reinforcement learning. The idea is to partition the state space into subsets, and define subtasks that perform transitions between the partitions. We formulate the problem of learning a hierarchical representation as an optimization problem that can be solved using gradient descent given a set of sampled trajectories. The resulting method can be applied to high-dimensional states (e.g.~images) and combined with state-of-the-art function approximation techniques for reinforcement learning.

In experiments, we show that our method can learn a useful subtask decompositions in several domains with high-dimensional observations in the form of images. We also show that the learned hierarchical representation can be used to transfer knowledge to new, previously unseen tasks, thus generalizing knowledge across tasks.

\subsection{Related Work}

Hierarchical reinforcement learning using hand-crafted subgoals to guide exploration, either as part of the value function representation~\cite{nachum2018near,schaul2015universal,sutton2017horde}, or as pseudo-rewards~\cite{eysenbach2019search,florensa2017automatic}, can solve hard exploration tasks with sparse rewards more efficiently than a flat learner, even for high-dimensional states.

Early work on learning hierarchical representations for reinforcement learning focused on analyzing the state transition graph~\cite{Menache02,Simsek05}, clustering nearby states~\cite{Mannor04}, discovering common substructure~\cite{Pickett02} or identifying landmarks~\cite{McGovern01,Simsek04,Solway14}. However, most of these methods rely on enumerating states, which is not possible in high-dimensional domains. \citeauthor{lakshminarayanan2016option}~(\citeyear{lakshminarayanan2016option}) used a spectral clustering algorithm on the state space representation learned using an unsupervised model prediction network~\cite{oh2015action}. Their method scales to high-dimensional states but strongly relies on the latent representation of the neural network and does not perform clustering directly on the original state space.

Skill learning~\cite{10.5555/1625275.1625420,10.5555/3042573.3042758} identifies initiation sets of options by searching backwards from a given set of target states where options terminate. Similar to our method, skills can be reused in a range of similar tasks. The option-critic framework~\cite{bacon2017option} and similar algorithms such as MODAC~\cite{abs-2102-06741} use gradient descent to learn the components of each option from trajectories. However, the resulting options are not easily interpretable, unlike our options that always transition between partitions.
DDO \cite{fox2017multi} leverages an Expectation-Maximization algorithm to train a hierarchy of options end-to-end for imitation learning. Unlike our method they do not explicitly consider initiation sets of the options.

\citeauthor{DBLP:conf/icml/CorneilGB18}~(\citeyear{DBLP:conf/icml/CorneilGB18}) learn a latent state model given a sequence of observations, akin to learning a mapping from states to abstract states, using a neural network architecture similar to a variational autoencoder.
\citeauthor{10.5555/3305890.3305918}~(\citeyear{10.5555/3305890.3305918}) use learned proto-value functions to identify subtask structure, in which a proto-value function becomes a local reward function for a given option.
\citeauthor{shang2019learning}~(\citeyear{shang2019learning}) use variational inference to construct a partition similar to ours. However, unlike our model-free option learning, the option policies are trained using dynamic programming, which requires knowledge of the environment dynamics. \citeauthor{eysenbach2019search}~(\citeyear{eysenbach2019search}) build distance estimates between pairs of states, and use the distance estimate to condition reinforcement learning in order to reach specific goals, which is similar to defining temporally extended actions.

Several authors have used state space partitions to handcraft hierarchical structure.
\citeauthor{ecoffet2019go}~(\citeyear{ecoffet2019go}) use a partition to learn to play Montezuma's revenge, using random search to find transitions between the partitions.
\citeauthor{NEURIPS2020_4a5cfa92}~(\citeyear{NEURIPS2020_4a5cfa92}) take advantage of equivalent partitions with the same local dynamics to reuse option policies in multiple partitions. When the number of termination states of options is relatively small, the resulting algorithm has much better sample complexity properties than flat learning.

\section{Background}

In this section we define Markov decision processes and hierarchical reinforcement learning in the form of the options framework. For any finite set $X$, let $\Delta(X)=\{p\in\mathbb{R}^X:\sum_{x\in X}p(x)=1, p(x)\geq 0 \; (\forall x)\}$ be the probability simplex on $X$, i.e.~the set of all probability distributions on $X$.

\subsection{Markov Decision Processes}

A finite Markov decision process (MDP) \cite{puterman2014markov} is a tuple $\mathcal{M} = \langle S,A,P,r \rangle$, where $S$ is a finite state space, $A$ is a finite action space, $P: S \times A \rightarrow \Delta(S)$ is a transition kernel and $r:S \times A \rightarrow \mathbb{R}$ is a reward function. At time $t$, the learning agent observes a state $s_t \in S$, takes an action $a_t \in A$, obtains a reward $r_t$ with expected value $\mathbb{E}[r_t] = r(s_t, a_t)$, and transitions to a new state $s_{t+1} \sim P(\cdot | s_t, a_t)$. We refer to $(s_t,a_t,r_t,s_{t+1})$ as a {\em transition}.

A stochastic policy $\pi:S\rightarrow\Delta(A)$ is a mapping from states to probability distributions over actions. The aim of reinforcement learning is to compute a policy $\pi$ that maximizes some notion of expected future reward. In this work, we consider the discounted reward criterion, for which the expected future reward of a policy $\pi$ can be represented using a value function $V^\pi$, defined for each state $s\in S$ as
\[
V^\pi(s) = \mathbb{E} \left[ \left. \sum_{t=1}^{\infty} \gamma^{t-1} r(S_t,A_t) \right\vert S_1=s \right].
\]
Here, random variables $S_t$ and $A_t$ model the state and action at time $t$, respectively, and the expectation is over the action $A_t\sim\pi(\cdot|S_t)$ and next state $S_{t+1}\sim P(\cdot|S_t,A_t)$. The discount factor $\gamma\in(0,1]$ is used to control the relative importance of future rewards, and to ensure $V^\pi$ is bounded.

As an alternative to the value function $V^\pi$, one can instead model expected future reward using an action-value function $Q^\pi$, defined for each state-action pair $(s,a)\in S\times A$ as
\[
Q^\pi(s, a) = \mathbb{E} \left[ \left. \sum_{t=1}^{\infty} \gamma^{t-1} r(S_t,A_t) \right\vert S_1=s, A_1=a \right].
\]
The value function $V^\pi$ and action-value function $Q^\pi$ are related through the well-known Bellman equations:
\begin{align*}
V^\pi(s) &= \sum_{a\in A} \pi(a|s) Q^\pi(s,a),\\
Q^\pi(s,a) &= r(s,a) + \gamma \sum_{s'\in S} P(s'|s,a) V^\pi(s').
\end{align*}
The aim of learning is to find an optimal policy $\pi^*$ that maximizes the value in each state, i.e.~$\pi^*(s) = \arg\max_\pi V^\pi$. The optimal value function $V^*$ and action-value function $Q^*$ satisfy the Bellman optimality equations:
\begin{align*}
 V^{*}(s) &= \max _{a \in A} Q^*(s,a),\\
Q^{*}(s, a) &= r(s,a)+ \gamma \sum_{s^{\prime} \in S} P(s^{\prime} | s, a) V^*(s').
\end{align*}

\subsection{Model-Based Methods}
If the reward function $r$ and transition probabilities $P$ are known (and the state and action spaces are not very large), we can use dynamic programming methods such as Value Iteration to compute an estimate $\widehat{Q}$ of the optimal action-value function, using the following update rule:

\begin{equation*}
\widehat{Q}_{t+1}(s, a) \gets r(s, a)+\gamma\sum_{s^{\prime} \in S} P(s^{\prime} | s, a) \max _{a^{\prime}} \widehat{Q}_t(s^{\prime}, a^{\prime}).
\end{equation*}
Value iteration repeatedly computes $\widehat{Q}_{t+1}$ for all state-action pairs $(s,a)$ until $\|\widehat{Q}_{t+1}-\widehat{Q}_{t}\|_\infty<\varepsilon$ for some desired accuracy $\varepsilon$, i.e.~until the difference between subsequent iterations is small enough.

\subsection{Function Approximation}

Since the state space $S$ is usually large, it is common to define a set of {\em features} $\Phi$, and an associated mapping $\phi:S \rightarrow \Phi$ from states to features. Value-based methods such as Deep Q-learning \cite{mnih2013playing} maintain an estimate $\widehat Q_\theta: \Phi\times A \rightarrow \mathbb{R}$ of the optimal action-value function, defined on features instead of states and parameterized on a vector $\theta$. Given a transition $\langle s_t,a_t,r_t,s_{t+1}\rangle$, DQN updates the parameter vector according to the gradient method:
\begin{equation*}
\theta \leftarrow \theta+\alpha\left(\operatorname{T}_{\widehat Q}-\widehat Q_\theta(s, a)\right) \nabla_{\theta} \widehat Q_\theta(s, a)
\end{equation*}

where $\operatorname{T}_{\widehat Q}$ is a target value based on optimal bellman equation calculated as:

\begin{equation*}
\operatorname{T}_{\widehat Q}=r\left(s, a\right)+\gamma \max _{a^{\prime}} \widehat Q_\theta\left(s^{\prime}, a^{\prime}\right)
\end{equation*}

\subsection{Options}

Given an MDP $\mathcal{M} = \langle S,A,P,r \rangle$, an option is a temporally extended action $o=\langle I^o, \pi^o, \beta^o \rangle$, where $I^o \subseteq S$ is an initiation set, $\pi^o: S \rightarrow \Delta(A)$ is a policy and $\beta^o: S \rightarrow [0, 1]$ is a termination function~\cite{sutton1999between}. An option can be applied in any state $s\in I^o$, selects actions using policy $\pi^o$, and terminates in a state $s'\in S$ with probability $\beta^o(s')$. A primitive action $a\in A$ is a special case of an option $\langle I^a, \pi^a, \beta^a \rangle$ with initiation set $I^a=S$, $\pi^a(a|s)=1$ and $\beta^a(s)=1 \; (\forall s)$, i.e.~the option can be applied in any state, terminates with probability $1$ in any state, and the associated policy always selects action $a$ with probability $1$.

Given a set of options $O$, we can form a semi-Markov decision process (SMDP) $\mathcal{S} = \langle S,O,P',R' \rangle$, where $P'$ and $R'$ model the transition probabilities and expected reward of options. Such an SMDP enables a learning agent to act and reason on multiple timescales. At the top level, the learning agent observes a state $s_t$, selects an option $o_t$, executes option $o_t$ until termination, and observes a next state $s_{t+k}$, where $k$ is the time it takes option $o_t$ to terminate. The reward $R_t=\sum_{u=t}^{t+k-1} \gamma^{u-t} r_u$ is the discounted sum of rewards obtained during the execution of option $o$. Hence $(s_t,o_t,R_t,s_{t+k})$ is a high-level transition, and with minor modifications, reinforcement learning algorithms can be adapted to estimate an optimal SMDP policy $\pi^*$, even when the SMDP dynamics $P'$ and $R'$ are unknown.

To learn the option policy $\pi^o$, it is common to define an option-specific reward function $r^o$, which defines an option-specific Markov decision process $\mathcal{M}^o = \langle S,A,P,r^o \rangle$. The policy $\pi^o$ is then implicitly defined as the optimal solution to $\mathcal{M}^o$. Even though the original definition of SMDPs considers options with fixed policies, in practice one can learn the option policies and the SMDP policy in parallel.

\section{Contribution}



In this section we present our main contribution, a method for learning a hierarchical representation of a given MDP.

\subsection{Compression Function}

The first step is to learn a compression function from MDP states to abstract states. We first define a set $Z$ of abstract states that will represent the partitions of the state space. WLOG, the elements of $Z$ are simply integers, i.e.~$Z=\{0,\ldots,|Z|-1\}$, where $|Z|$ is an input parameter of the method. Our goal is to learn a parameterized compression function $f_\psi:S\rightarrow\Delta(Z)$ that maps MDP states to probability distributions over abstract states. Ideally, $f_\psi$ should be {\em deterministic}, but the learning framework we consider favors probabilistic compression functions.

Intuitively, for abstract states to represent partitions of the state space, on a given trajectory the abstract state should remain the same most of the time, and only change occasionally. We formalize this intuition as a loss term, which will later be part of the objective that the learner attempts to minimize. Let $\tau = \langle s_t,a_t,r_t,s_{t+1}\rangle$ be a transition, and let $\mathcal{T}=\{\tau_1,\ldots,\tau_m\}$ be a set of transitions. The loss associated with $\mathcal{T}$ is given by
\[
\mathcal{L}_Z(\mathcal{T}) =- \sum_{\tau\in\mathcal{T}} \sum_{z\in Z} f_\psi(z|s_t) \log f_\psi(z|s_{t+1}).
\]
Here, $-\sum_{z\in Z}f_\psi(z|s_t) \log f_\psi(z|s_{t+1})$ is the cross-entropy loss for consecutive states $s_t$ and $s_{t+1}$ in $\tau$, measuring the distance between the distributions $f_\psi(\cdot|s_t)$ and $f_\psi(\cdot|s_{t+1})$.

On its own, the above loss term will not yield a meaningful compression function, since it can be minimized by mapping all states to the same abstract state. To ensure that all abstract states appear in the compression, we define a second loss term equivalent to the negative entropy of the compression function across the same set of transitions $\mathcal{T}$. Given an abstract state $z\in Z$, let $F(z|\mathcal{T}) = \frac 1 {|\mathcal{T}|} \sum_{\tau\in \mathcal{T}} f_\psi(z|s_t)$ be the average probability of being in $z$ across the first state $s_t$ of each transition $\tau\in\mathcal{T}$. We define a loss term
\[
\mathcal{L}_H(\mathcal{T}) = - H(F(\cdot|\mathcal{T})) = \sum_{z\in Z} F(z|\mathcal{T}) \log F(z|\mathcal{T}),
\]
where $H(F(\cdot|\mathcal{T}))$ is the entropy of the function $F(\cdot|\mathcal{T})$. This loss is minimized when the probabilities of abstract states are uniform, i.e.~each abstract state is equally likely.

Finally, as already stated, we would like the compression function $f_\psi$ to be as deterministic as possible. For this reason, we define a third loss term equivalent to the entropy of the compression function for individual states. We use the same set of transitions $\mathcal{T}$, and define this loss term as
\begin{align*}
\mathcal{L}_D(\mathcal{T}) &= \frac 1 {|\mathcal{T}|} \sum_{\tau\in \mathcal{T}} H(f_\psi(\cdot|s_t))\\
&= - \frac 1 {|\mathcal{T}|} \sum_{\tau\in \mathcal{T}} \sum_{z\in Z} f_\psi(z|s_t) \log f_\psi(z|s_t).
\end{align*}
This loss term is minimized when the compression function $f_\psi(\cdot|s_t)$ is deterministic, i.e.~assigns probability $1$ to a single abstract state, for the first state $s_t$ of each transition $\tau\in\mathcal{T}$.

The overall loss function $\mathcal{L}(\mathcal{T})$ is a combination of the three individual loss terms, i.e.
\begin{equation}\label{eq:loss}
\mathcal{L}(\mathcal{T}) = \mathcal{L}_Z(\mathcal{T}) + w_H \mathcal{L}_H(\mathcal{T}) + w_D \mathcal{L}_D(\mathcal{T}),
\end{equation}
where $w_H$ and $w_D$ are weights that we can tune to determine the relative importance of each loss term. Note that for $w_D=-1$, $\mathcal{L}_Z(\mathcal{T}) + w_D \mathcal{L}_D(\mathcal{T})$ is the average Kullback-Leibler divergence between $f_\psi(\cdot|s_t)$ and $f_\psi(\cdot|s_{t+1})$; however, our intention is to use positive values of $w_D$.

To train our compression function we use experience replay~\cite{mnih2013playing} to randomize the transitions in $\mathcal{T}$. We first sample a set of trajectories using some exploration policy, which constitutes our memory. However, learning directly from consecutive transitions along the same trajectory is inefficient, due to the strong correlations between the samples. Instead, we form the set of transitions $\mathcal{T}$ by randomly sampling individual transitions from the memory. Randomizing the sampled transitions this way breaks the correlations and therefore reduces the variance of the updates.

\subsection{Hierarchical Representation}

Once we have learned a compression function $f_\psi$ for a given MDP $\mathcal{M} = \langle S,A,P,r \rangle$, we use it to define a set of options $O$ and an SMDP $\mathcal{S}$. First, we introduce a {\em deterministic} compression function $g:S\rightarrow Z$, defined in each state $s$ as $g(s)=\arg\max_z f_\psi(z|s)$. Given an abstract state $z\in Z$, let $S_z$ be the subset of states that map to $z$, i.e.~$S_z=\{s\in S:g(s)=z\}$. 

Our algorithm then uses the compression function in an online manner, by exploring the environment and finding {\em abstract transitions}, i.e.~consecutive states $s_t$ and $s_{t+1}$ such that $g(s_t)\neq g(s_{t+1})$. Let $\mathcal{Y}\subseteq Z\times Z$ be the subset of pairs of distinct abstract states $(z,z')$ that appear as abstract transitions while exploring, i.e.~there exist two consecutive states $s_t$ and $s_{t+1}$ such that $g(s_t)=z$ and $g(s_{t+1})=z'$. For each pair $(z,z')\in\mathcal{Y}$, we introduce an option $o_{z,z'}=\langle I^o,\pi^o,\beta^o \rangle$ whose purpose is to perform an abstract transition from $z$ to $z'$. Option $o_{z,z'}$ is applicable in abstract state $z$, i.e.~$I^o=S_z$, and terminates as soon as we reach an abstract state different from $z$, i.e.~$\beta^o(s)=0$ if $s\in S_z$ and $\beta^o(s)=1$ otherwise.

To learn the policy $\pi^o$ of option $o_{z,z'}$, we define an option-specific Markov decision process $\mathcal{M}^o = \langle S_z,A,P,r^o \rangle$. Note that $\mathcal{M}^o$ needs only be defined for states in $S_z$, since option $o_{z,z'}$ always terminates outside this set. The local reward function $r^o$ is defined for each state-action pair as $r^o(s,a)=r(s,a)$, i.e.~equal to the environment reward. We also introduce a bonus $+1$ for terminating in a state $s$ such that $g(s)=z'$. As a consequence, the policy $\pi^o$ has an incentive to leave abstract state $z$, and prefers to transition to abstract state $z'$ whenever possible.

Let $O=\{o_{z,z'}:(z,z')\in\mathcal{Y}\}$ be the set of options for performing abstract transitions, and let $O_z=\{o\in O:I^o=S_z\}$ be the subset of options applicable in abstract state $z$. We define an SMDP $\mathcal{S} = \langle S,O,P',R' \rangle$, i.e.~the high-level choices of the learning agent are to select abstract transitions to perform. Once the individual option policies have been trained, exploration is typically more efficient since the single decision of which option to execute results in a state that is many steps away from the initial state. In addition, one can approximate the SMDP policy as $\pi:Z\rightarrow\Delta(O)$, i.e.~the choice of which option to execute only depends on the current abstract state. This has the potential to significantly speed up learning if $|Z|\ll|S|$.

The system is trained using a Manager-Worker architecture \cite{feudal}. The Manager performs tabular Value Iteration over the SMDP. The motivation for using tabular learning is that the number of abstract states $|Z|$ is typically small, even if states are high-dimensional. On the other hand, the Worker uses off-policy value based methods to learn the policies of the options $o_{z, z'}$.

\subsection{Controllability} \label{subsec: Controllability}

According to the definition of the option reward function $r^o$, option $o_{z,z'}$ is equally rewarded for reaching any boundary state between abstract states $z$ and $z'$. However, all boundary states may not be equally valuable, i.e.~from some boundary states, the options in $O_{z'}$ may have a higher chance of terminating successfully. To encourage option $o_{z,z'}$ to reach valuable boundary states and thus make the algorithm more robust to the choice of compression function $g$, we add a reward bonus when the option successfully terminates in a state $s'$ belonging to abstract state $z'$.

One possibility is that the reward bonus depends on the value of state $s'$ of options in the set $O_{z'}$. However, this introduces a strong coupling between options in the set $O$: the value function $V_{z,z'}$ of option $o_{z,z'}$ will depend on the value functions of options in $O_{z'}$, which in turn depend on the value functions of options in neighboring abstract states of $z'$, etc. We want to avoid such a strong coupling since learning the option value functions may become as hard as learning a value function for the original state space $S$.

Instead, we introduce a reward bonus which is a proxy for controllability, by counting the number of successful applications of subsequent options after $o_{z,z'}$ terminates. Let $M$ be the number of options that are selected after $o_{z,z'}$, and let $N\leq M$ be the number of such options that terminate successfully. We define a controllability coefficient $\rho$ as
\begin{equation}
	\rho(z) = \frac{N}{M}.
\end{equation}
We then define a modified reward function $\bar{r}^o$ which equals $r^o$ except when $o_{z,z'}$ terminates successfully, i.e.~$\bar{r}^o(s,a,s') = r^o(s,a,s') + \rho(z)$ if $s'\in z'$. In experiments we use a fixed horizon $M=4$ after which we consider successful option transitions as irrelevant.

\subsection{Transfer}

The hierarchical representation in the form of the SMDP $\mathcal{S}$ defined above can be used to transfer knowledge between tasks. Concretely, we assume that the given MDP $\mathcal{M}$ can be extended to form a task by adding states and actions. Imagine for example that $\mathcal{M}$ models a navigation problem in a given environment. A task can be defined by adding objects in the environment that the learning agent can manipulate, while navigation is still part of the task.

Formally, given an MDP $\mathcal{M} = \langle S,A,P,r \rangle$, a task $\mathbb{T}$ is an MDP $\mathcal{M}_\mathbb{T} = \langle S\times S_\mathbb{T}, A \cup A_\mathbb{T}, P \cup P_\mathbb{T}, r \cup r_\mathbb{T} \rangle$. The states in $S_\mathbb{T}$ represent information about task-specific objects, and the actions in $A_\mathbb{T}$ are used to manipulate these objects. The transition kernel $P_\mathbb{T}:(S \times S_\mathbb{T}) \times A_\mathbb{T} \rightarrow \Delta(S_\mathbb{T})$ governs the effects of the actions in $A_\mathbb{T}$, which may depend on the states of the original MDP (e.g.~the location of the agent). Finally, the reward function $r_\mathbb{T}:(S \times S_\mathbb{T}) \times A_\mathbb{T} \rightarrow \mathbb{R}$ models the reward associated with actions in $A_\mathbb{T}$.

To solve a task, we can replace the MDP $\mathcal{M}$ with the learned SMDP $\mathcal{S} = \langle S,O,P',R' \rangle$, forming a task SMDP $\mathcal{S}_\mathbb{T} = \langle S\times S_\mathbb{T}, O \cup A_\mathbb{T}, P' \cup P_\mathbb{T}, R' \cup r_\mathbb{T} \rangle$. Here, the options in $O$ are used to navigate in the original state space $S$, while the actions in $A_\mathbb{T}$ are used to manipulate the task-specific objects. If the policies of the options in $O$ have been previously trained, the task SMDP $\mathcal{S}_\mathbb{T}$ can significantly accelerate learning compared to the task MDP $\mathcal{M}_\mathbb{T}$. To ensure that the learning agent can navigate to individual objects inside a partition of $Z$, we consider states in $S_\mathbb{T}$ to be different abstract states; hence our algorithm will automatically add options for manipulating objects.

\begin{figure}[t]
    \centering
    \includegraphics[width=3cm]{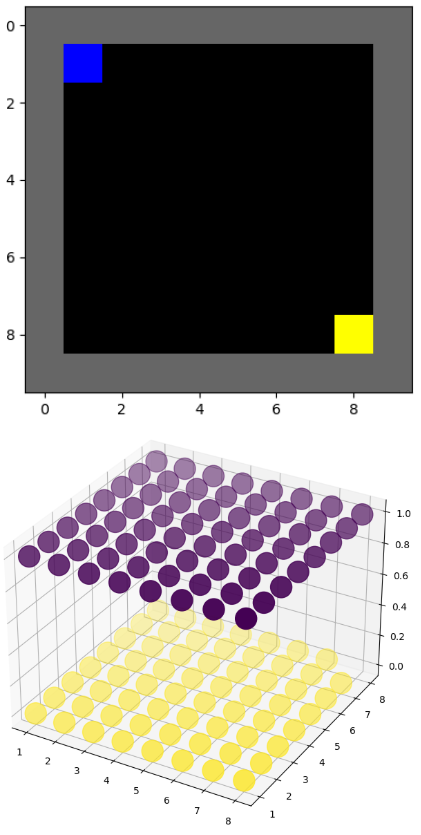}
    \caption{Results on Key-Door0 gridworld environment. The first row represents the environments, and the second row illustrates the corresponding learned deterministic compression functions, where different colors represent different abstract states $z\in Z$.}%
    \label{fig:compressionKeyDoor}%
\end{figure}

\begin{figure*}[t]
    \centering
    \includegraphics[width=16.5cm]{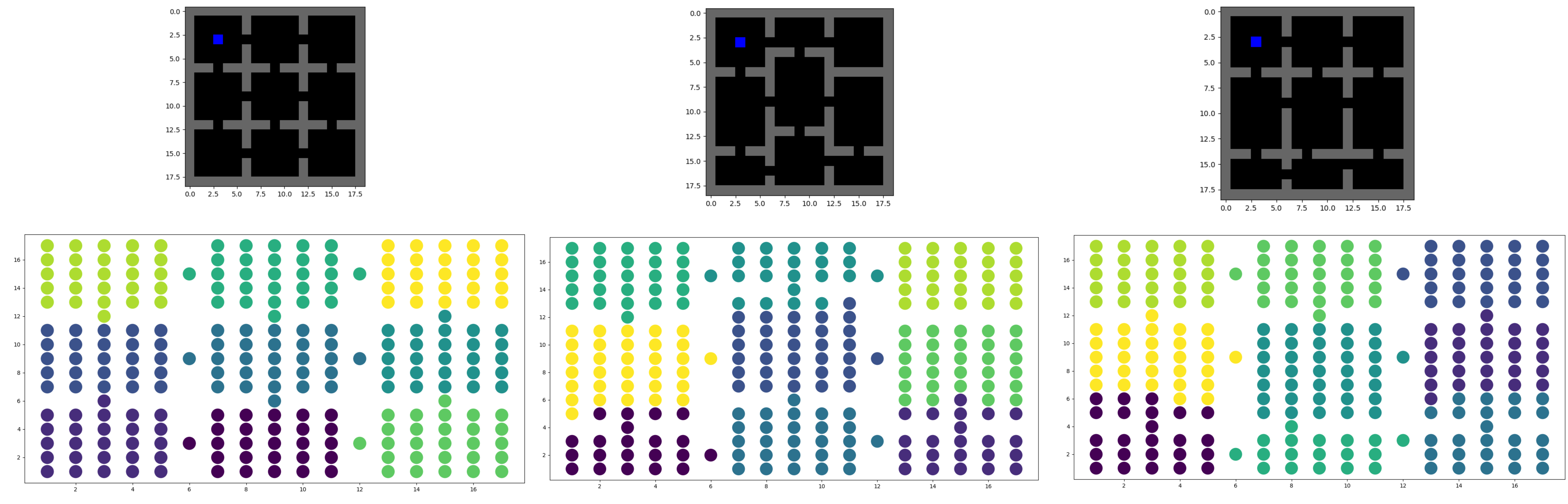}
    \caption{Results on geometric variations of the NineRooms0 gridworld environments. The first row represents the environments, and the second row illustrates examples of the corresponding learned deterministic compression functions, where different colors represent different abstract states $z\in Z$.}%
    \label{fig:compression}%
\end{figure*}

\section{Experimental Results}

The experiments are designed to answer the following questions:

\begin{itemize}
    \item Is the learned compression function suitable for learning a hierarchy?
    \item Does the learned hierarchy transfer across different tasks in the same environment?
    \item How does our HRL algorithm compare against state-of-the-art flat algorithms, such as Self Imitation Learning~\cite{oh2018self} and Double-DQN with Prioritized Experience Replay~\cite{schaul2016prioritized}? 
\end{itemize}



\noindent
Subject to acceptance of the paper, we plan to make the code publicly available to reproduce the experimental results.

\subsection{Learning a Compression Function}

We designed two different empty navigation environments without tasks, KeyDoor0 (c.f.~Figure~\ref{fig:compressionKeyDoor}), with grid size $10\times 10$, where an agent (blue square) always starts in position (1, 1), has to collect a key (yellow square). And NineRooms0 (c.f.~Figure~\ref{fig:compression}), a nine rooms grid environment with grid size $19\times 19$ where at each episode the agent is placed at a random initial position, which promotes exploration. 
For all the environments the states are $(x,y)$-positions which are mapped to images, and the discrete action space is $A = \{ up, down, left, right \}$.

The first step of our procedure consists in a pre-training phase where we form a replay memory of trajectories. We use a random exploration policy to repeatedly generate trajectories from the random initial states, using a fixed episode length of 100. During this phase, we can vary the number of trajectories generated to test the robustness of the approach.

We then use the replay memory and a number of abstract states $|Z|=2$ for KeyDoor0 and $|Z|=9$ for NineRoom0 to train the compression function $f_\phi$ using the AdamW optimizer~\cite{loshchilov2017decoupled} by minimizing the loss in~\eqref{eq:loss} over 4000 iterations, randomly sampling a set of 32 transitions $\mathcal{T}$ from the replay memory in each iteration. The learned compression functions for 1000 trajectories are shown in Figures~\ref{fig:compressionKeyDoor} and~\ref{fig:compression}, respectively.

To asses the robustness of our procedure, in Figure~\ref{fig:compression} we evaluate how the compression function changes by introducing different geometries of the NineRooms0 environment. As we can see, if we make the room sizes imbalanced, the resulting compression function does not exactly match the shape of the rooms, due to the second term $\mathcal{L}_H$ of the loss in~\eqref{eq:loss}, which promotes all abstract states $z$ to be equally likely. However, the resulting compression function still partitions the states and translates into a correct SMDP.

\begin{figure}[t]
    \centering
    \includegraphics[width=9cm]{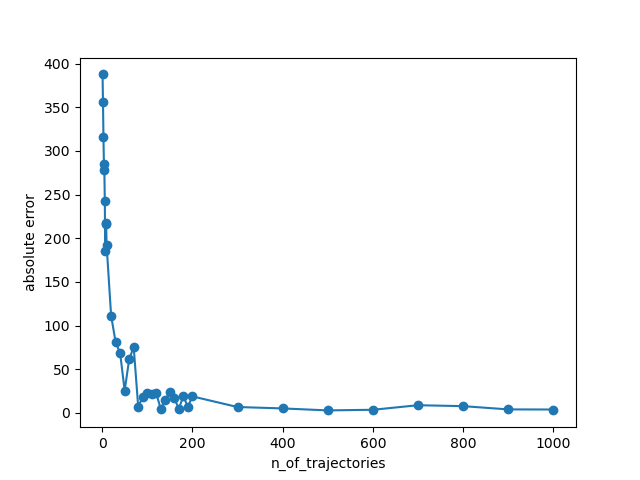}
    \caption{Absolute error of the compression function, evaluated on increasing replay memory size.}
    \label{fig:ntraj_abserror}
\end{figure}

In Figure~\ref{fig:ntraj_abserror}, we evaluate how the size of the replay memory affects the accuracy of the compression function in terms of the absolute error deviation with respect to a correct representation. For this experiment we use the left-most room in Figure~\ref{fig:compression} with balanced room sizes, and vary the number of trajectories in the replay memory. When the replay memory contains at least $200$ trajectories, the procedure converges to an absolute error very close to 0, while less than $200$ trajectories results in an increasing absolute error.

In the supplementary material we present additional experiments with learning a compression function in the MountainCar environment. We also list all hyperparameters of the algorithm.

\subsection{Hierarchical Reinforcement Learning}

Following the pre-training phase, we can use the learned compression function to solve any task in the same environment. In what follows we use the compression function learned on the left-most room in Figure~\ref{fig:compression} with a replay memory of 1000 trajectories.
We distinguish between a {\em manager} in charge of solving the task SMDP $\mathcal{S}_\mathbb{T}$ and {\em workers} in charge of solving the option MDPs $\mathcal{M}^o$.

\subsubsection{Manager}

Our algorithm iteratively grows an estimate of the SMDP $S$. Initially, the agent only observes a single state $s \in S$ and associated abstract state $z = g(s)$. Hence the state space $Z$ contains a single abstract state $z$, whose associated option set $O_z$ is initially empty. In this case, the only alternative available to the agent is to $explore$. For each abstract state $z$, we add an exploration option $o_z^{exploration} = \langle z, \pi_z^{exploration}, \beta_z \rangle$ to the option set $O$. This option has the same initiation set and termination condition as the options in $O_z$, but the policy $\pi_z^{exploration}$ is an exploration policy that selects actions uniformly at random, terminating when it leaves abstract state $z$ or exhausts a given budget.

\begin{figure}[t]
    \centering
    \includegraphics[width=5cm]{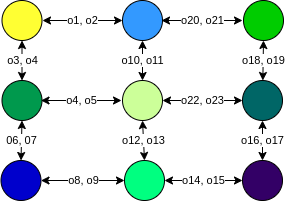}
    \caption{The discovered invariant SMDP.}
    \label{fig:SMDP}
\end{figure}

Once the agent discovers a neighboring abstract state $z'$ of $z$, it adds $z'$ to the set $Z$ and the associated option $o_{z, z'}$ to the option set $O$. The agent also maintains and updates a directed graph whose nodes are abstract states and whose edges represent the neighbor relation. Hence next time the agent visits abstract state $z$, one of its available actions is to select option $o_{z, z'}$. When option $o_{z,z'}$ is selected, it chooses actions using its policy $\pi_{z,z'}$ and updates $\pi_{z,z'}$ based on the rewards of the option MDP $\mathcal{M}^o$. Figure~\ref{fig:SMDP} shows an example representation discovered by the algorithm.

Algorithm \ref{algo: manager} shows the pseudo-code of the algorithm. As explained, $Z$ is initialized with the abstract state $z$ of the initial state $s$, and $O$ is initialized with the exploration option $o_z^{exploration}$. In each iteration the algorithm selects an option $o$ which is applicable in the current abstract state $z$. If we transition to a new abstract state $z'$, it is added to $Z$ and the exploration option $o_{z'}^{exploration}$ and transition option $o_{z,z'}$ are appended to $O$. The process then repeats from the next state $s'$.

The subroutine $\textsc{GetOption}$ that selects an option $o$ in the current abstract state $z$ can be implemented in different ways; in our case, we use an $\epsilon$-greeedy policy.



Since the set of abstract states $Z$ is small, the manager performs tabular Value Iteration over the task SMDP $\mathcal{S}_\mathbb{T}$.

In order to recognize new goal states, while exploring we define any terminal state in the environment as a new abstract state $z$; hence the manager will introduce options for reaching this terminal state.


\subsubsection{Planning with a learned SMDP}

We have seen how the state space of the task SMDP $\mathcal{S}_\mathbb{T}$ is discovered online given a compression function $g(s)$. 
In order to apply a model-based method on this learned compression function, we still need to be able to estimate the transition kernel $P_{\mathcal{S}_\mathbb{T}}$ and reward function $r_{\mathcal{S}_\mathbb{T}}$ of $\mathcal{S}_\mathbb{T}$.

Estimating the transition probability associated with an option $o_{z,z'}$ of our task SMDP
is not easy, since the policy $\pi^o$ is trained online while exploring the environment, making the transition probability non-stationary. In order to alleviate the cost of estimating the transition probability, we assume that $o_{z,z'}$ will become deterministic once the training phase terminates, i.e.~$\widehat{P}_{\mathcal{S}_\mathbb{T}}( z' | z, o_{z,z'}) = 1$. Though this is an approximation, the aim of option $o_{z,z'}$ is precisely to reach abstract state $z'$, and constructing the SMDP is intended to simplify the high-level decision making.

On the other hand, for each state-option pair $(z,o)$ of the task SMDP $\mathcal{S}_\mathbb{T}$, we estimated the task SMDP reward $r_{\mathcal{S}_\mathbb{T}}$ as an average of the reward encountered in the environment:
\begin{equation*}
    \widehat{r}_{\mathcal{S}_\mathbb{T}}(s, o) = \frac{\sum_{i=1}^{N(z,o)} R_i(z, o)}{N(z, o)},
\end{equation*}
where $N(z,o)$ counts the number of times the state-option pair $(z,o)$ has been observed, and $R_i$ is the cumulative reward obtained while applying option $o$ for the $i$-th time.

Since the model changes over time, the subroutine $\textsc{UpdatePolicy}$ updates the $Q$ values of the Manager at regular intervals by applying value iteration on the learned SMDP $\widehat{\mathcal{S}}_\mathbb{T}$.

\begin{figure}[t]
    \centering
    \includegraphics[width=5cm]{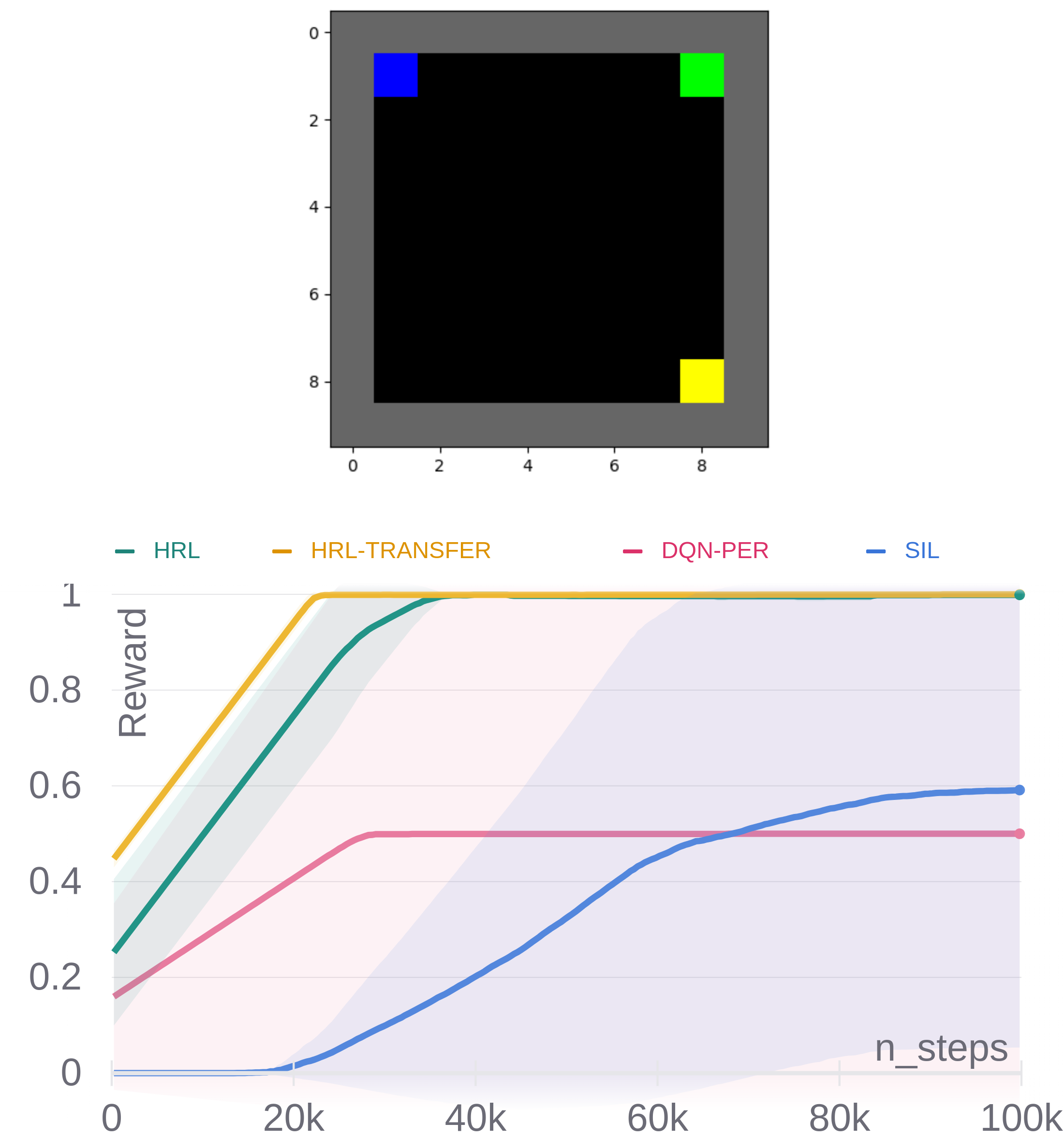}
    \caption{Results in the KeyDoor environment.}%
    \label{fig:results KeyDoor}%
\end{figure}

\subsubsection{Workers}
The workers are in charge of learning the policies of each option $o_{z, z'}$ in $O$, allowing the manager to transition between two abstract states $z, z'$. We use Double DQN \cite{vanhasselt2015deep}, a version of DQN that addresses the overestimation of $Q$-values, combined with Prioritized Experience Replay (PER) \cite{schaul2016prioritized} that improves the way experience is sampled from the Experience Replay. The rewards that the worker observes are defined in the Hierarchical Representation Section, and implemented in the routine $\textsc{TrainOption}$ from Algorithm~\ref{algo: manager}.

Since Double DQN is able to evaluate Q-values off-policy, one can relabel failed transitions to speed up learning of the correct option behavior, similar to Hindsight Experience Replay~\cite{andrychowicz2017hindsight}. The architecture is made of a neural network $Q_\theta$ parametrized on $\theta$, and a frozen target network $Q_{\bar{\theta}}$ used to alleviate the non-stationarity of the targets $\operatorname{T_Q} = r(s, a) + \gamma \, \max_{a'} Q_\theta(s', a')$.

The parameters of the neural network are updated as: 
\begin{equation*}
\theta \leftarrow \theta+\alpha\left(\operatorname{T}_{\mathrm{D}}-Q_\theta(s, a)\right) \nabla_{\theta} Q_\theta(s, a),
\end{equation*}
where $\operatorname{T}_{\mathrm{D}}$ is the target value computed as:
\begin{equation*}
\begin{split}
\operatorname{T}_{\mathrm{D}}=r\left(s, a\right)+ & \gamma Q_{\bar{\theta}}\left(s_{t+1}, \underset{a}{\operatorname{argmax}} Q_{\theta}\left(s_{t+1}, a\right) \right).
\end{split}
\end{equation*}
The target network is then updated with Polyak updates~\cite{heess2015learning}:
\begin{equation*}
   \bar{\theta} = \tau \, \theta + (1-\tau) \, \bar{\theta}.
\end{equation*}

        \begin{algorithm}[H]
		\caption{\textsc{Manager}}
		\large
		\begin{algorithmic}[1]
			 \STATE {\bf Input}: environment $e$, previously discovered SMDP $\mathcal{S}$ in case of transfer learning, compression function $g$
			 \STATE $s \gets initial state$
			 \STATE $z \gets g(s)$
			 \IF{$z \not \in Z$}
			 \STATE $Z \gets Z \cup \{z\}$
			 \STATE $O \gets {o_z^{exploration}}$
			 \ENDIF
			 \STATE $\pi_{\mathbb{T}} \gets $ initial policy
			 \STATE $o \gets$ None
			 \WHILE{within budget}
			    \IF {$o$ is None or Terminate}
				 \STATE $o  \gets \textsc{GetOption}(\pi_{\mathbb{T}}, z,O)$
				 \STATE $R = 0$
				 \ENDIF
				 \STATE $s',r, done \gets e(o(s))$
				 \STATE $\textsc{TrainOption}(o, s, r, s', done)$
				 \STATE $R = R + r$
				 \STATE $z' \gets g(s')$
				 \IF{$z' \not \in Z$}
				 \STATE $Z \gets Z \cup \{z'\}$
				 \STATE $O \gets O \cup \{o_{z'}^{exploration}, o_z^{z, z'}\}$
				 \ENDIF
				 \IF{$z \neq z'$}
				    \STATE $\textsc{UpdatePolicy}(\pi_{\mathbb{T}}, z, o, R, z')$
				    \STATE $o  \gets \textsc{GetOption}(\pi_{\mathbb{T}}, z',O)$
				 \ENDIF
				 \IF{$s'$ is terminal and $s'$ not in $Z$}
				    \STATE $O \gets O \cup \{o_z^{z,s'}\}$
				    \STATE $Z \gets Z \cup \{s'\}$
				 \ENDIF
				 \STATE $(z,s) \gets (z',s')$
			\ENDWHILE
		 \end{algorithmic}
		 \label{algo: manager}
		\end{algorithm}

\begin{figure*}[t]
    \centering
    \includegraphics[width=16cm]{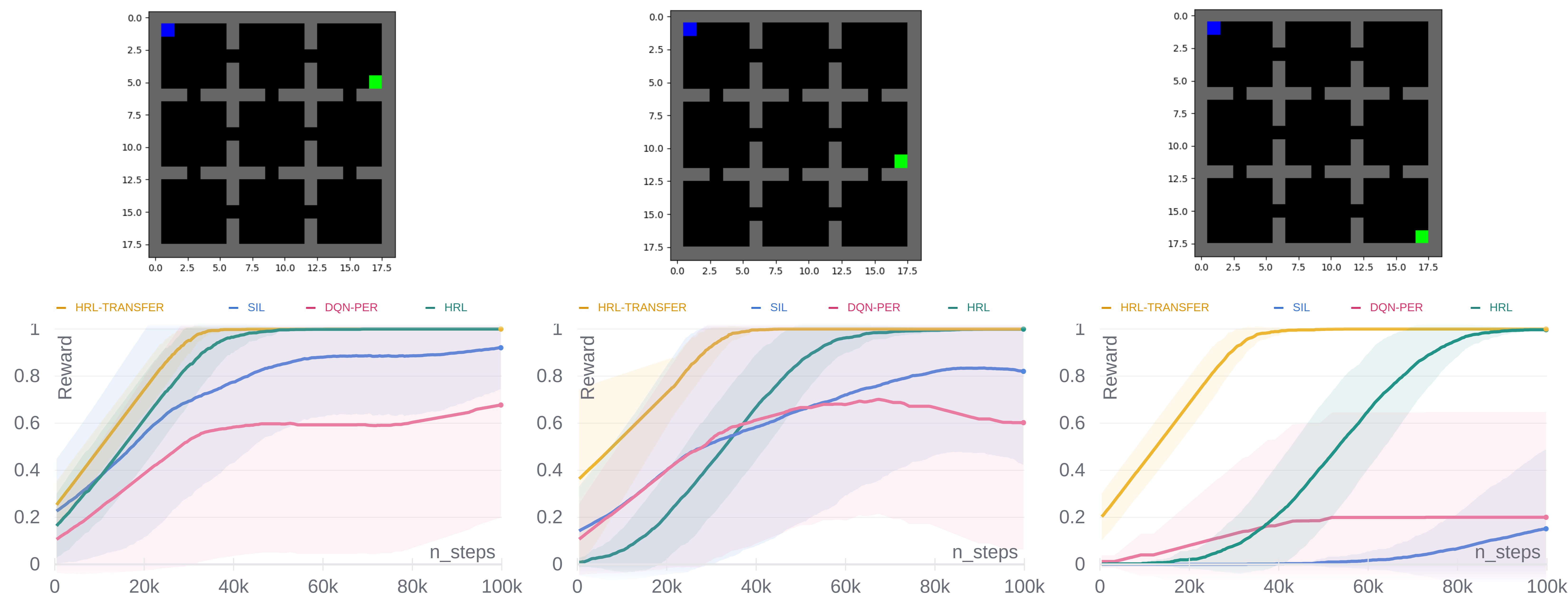}
    \caption{ Results on the variations of nine room gridworld environments where the goal (green square) is placed at an increasing distance from the agent (blue square). From left to right: NineRoom1, NineRoom2, NineRoom3.}%
    \label{fig: results}%
\end{figure*}

\subsection{Experiments}

In our experiments, we evaluate the performance of our agent in two environments, on a KeyDoor environment in Figure ~\ref{fig:results KeyDoor} where an agent has to collect a key (yellow square) and open a door (green square) and a NineRooms environment in Figure ~\ref{fig: results} where an agent has to reach the goal (green square). In both environments the initial position is fixed to $(1,1)$. 
In the NineRooms environment we defined three variants where the goal is positioned at an increasing distance from the initial state to make exploration harder, c.f.~NineRooms1, NineRooms2 and NineRooms3 in Figure~\ref{fig: results}. 
Results are averaged over 5 seeds and each experiment is run for 100{,}000 iterations. Even though the compression function is given, the goal location is unknown, so the agent has to explore the environment in order to find the goal location for the first time.

We set the maximum number of steps in the environment to 40 for the KeyDoor environment and to 200 for NineRooms environment, making exploration hard, especially in NineRooms3 where the goal is at the maximum distance from the initial state. Results in Figures~\ref{fig:results KeyDoor} and~\ref{fig: results} show the total reward with a running average smoothing of 100 episodes and shaded standard deviation. In the KeyDoor environment the agent receives a reward of +1 only once it opens the door (green square) with the key (yellow square) while in NineRooms the agent receives a reward of +1 when it reaches the goal position (green square) and a reward of 0 elsewhere.

We compared our algorithm against state of the art flat reinforcement learning agents designed to perform well in sparse reward settings, namely Self Imitation Learning (SIL) \cite{oh2018self} and Double DQN with Prioritized Experience Replay (DQN-PER) \cite{schaul2016prioritized}, implemented on top of the Reinforcement Learning framework Machin~\cite{machin}.

We refer to "HRL" as our algorithm in which the task SMDP $\mathcal{S}_{\mathbb{T}}$ has to be learned online while exploring, but the compression function $g$ is given. "HRL-TRANSFER" refers to our algorithm where the agent is first pretrained in order to learn the options on NineRooms0 in Figure~\ref{fig:compressionKeyDoor} without any task and then exposed to NineRooms1, NineRooms2, NineRooms3 in sequence. In this case the algorithm benefits from the transfer of the SMDP $S$ while the manager policy (i.e.~Q-values) are reset to 0 after training in each environment.

We can observe that the HRL algorithm learns faster than SIL and DQN-PER in all the environments. SIL and DQN-PER both rely only on random exploration, but once they find a positive reward, they can exploit it. In contrast, the exploration of HRL and HRL-TRANSFER are aided by the hierarchical structure. Both SIL and DQN-PER present high variance, and for some seeds they are not even able to solve the task, given the budget of 100{,}000 iterations. We can also observe that HRL-TRANSFER does improve over HRL, and we would argue that this improvement could be larger if we choose harder tasks were the option policies for transitioning between abstract states become harder to learn.

\section{Conclusion}

We present a novel method for learning a hierarchical representation from sampled transitions in high-dimensional domains. The idea is to generate abstract states that partition the original state space, and introduce options for performing transitions between abstract states. Experiments show that the learned representation can successfully be used to solve multiple tasks in the same environment, significantly speeding up learning compared to a flat learner.

An important direction for future work is to sample trajectories using a more informed exploration policy, since learning the compression function depends on having a variety of trajectories in different states. Another possible extension is to interleave representation learning with policy improvement, which may successively improve the quality of the sampled trajectories. Yet another possibility is to correct the compression function in states from which some abstract transitions are not possible.

%
%
%


\bibliography{reference}

\begin{thebibliography}{36}
\providecommand{\natexlab}[1]{#1}

\bibitem[{Andrychowicz et~al.(2017)Andrychowicz, Wolski, Ray, Schneider, Fong,
  Welinder, McGrew, Tobin, Abbeel, and Zaremba}]{andrychowicz2017hindsight}
Andrychowicz, M.; Wolski, F.; Ray, A.; Schneider, J.; Fong, R.; Welinder, P.;
  McGrew, B.; Tobin, J.; Abbeel, O.~P.; and Zaremba, W. 2017.
\newblock Hindsight experience replay.
\newblock In \emph{Advances in Neural Information Processing Systems},
  5048--5058.

\bibitem[{Bacon, Harb, and Precup(2017)}]{bacon2017option}
Bacon, P.-L.; Harb, J.; and Precup, D. 2017.
\newblock The option-critic architecture.
\newblock In \emph{Proceedings of the 31st AAAI Conference on Artificial
  Intelligence}.

\bibitem[{Barto and Mahadevan(2003)}]{10.1023/A:1022140919877}
Barto, A.~G.; and Mahadevan, S. 2003.
\newblock Recent Advances in Hierarchical Reinforcement Learning.
\newblock \emph{Discrete Event Dynamic Systems}, 13(1–2): 41–77.

\bibitem[{Corneil, Gerstner, and Brea(2018)}]{DBLP:conf/icml/CorneilGB18}
Corneil, D.~S.; Gerstner, W.; and Brea, J. 2018.
\newblock Efficient ModelBased Deep Reinforcement Learning with Variational
  State Tabulation.
\newblock In \emph{Proceedings of the 35th International Conference on Machine
  Learning, {ICML} 2018, Stockholmsm{\"{a}}ssan, Stockholm, Sweden, July 10-15,
  2018}, 1057--1066.

\bibitem[{\c{S}im\c{s}ek and Barto(2004)}]{Simsek04}
\c{S}im\c{s}ek, {\"O}.; and Barto, A. 2004.
\newblock Using relative novelty to identify useful temporal abstractions in
  reinforcement learning.
\newblock \emph{Proceedings of the International Conference on Machine
  Learning}, 21: 751--758.

\bibitem[{\c{S}im\c{s}ek, Wolfe, and Barto(2005)}]{Simsek05}
\c{S}im\c{s}ek, {\"O}.; Wolfe, A.; and Barto, A. 2005.
\newblock Identifying useful subgoals in reinforcement learning by local graph
  partitioning.
\newblock \emph{Proceedings of the International Conference on Machine
  Learning}, 22.

\bibitem[{Da~Silva, Konidaris, and Barto(2012)}]{10.5555/3042573.3042758}
Da~Silva, B.~C.; Konidaris, G.; and Barto, A.~G. 2012.
\newblock Learning Parameterized Skills.
\newblock In \emph{Proceedings of the 29th International Coference on
  International Conference on Machine Learning}, ICML'12, 1443–1450. Madison,
  WI, USA: Omnipress.
\newblock ISBN 9781450312851.

\bibitem[{Dayan and Hinton(2000)}]{feudal}
Dayan, P.; and Hinton, G. 2000.
\newblock Feudal Reinforcement Learning.
\newblock \emph{Advances in Neural Information Processing Systems}, 5.

\bibitem[{Ecoffet et~al.(2019)Ecoffet, Huizinga, Lehman, Stanley, and
  Clune}]{ecoffet2019go}
Ecoffet, A.; Huizinga, J.; Lehman, J.; Stanley, K.~O.; and Clune, J. 2019.
\newblock Go-explore: a new approach for hard-exploration problems.
\newblock \emph{arXiv preprint arXiv:1901.10995}.

\bibitem[{Eysenbach, Salakhutdinov, and Levine(2019)}]{eysenbach2019search}
Eysenbach, B.; Salakhutdinov, R.~R.; and Levine, S. 2019.
\newblock {Search on the replay buffer: Bridging planning and reinforcement
  learning}.
\newblock In \emph{Advances in Neural Information Processing Systems},
  15220--15231.

\bibitem[{Florensa et~al.(2017)Florensa, Held, Geng, and
  Abbeel}]{florensa2017automatic}
Florensa, C.; Held, D.; Geng, X.; and Abbeel, P. 2017.
\newblock Automatic goal generation for reinforcement learning agents.
\newblock \emph{arXiv preprint arXiv:1705.06366}.

\bibitem[{Fox et~al.(2017)Fox, Krishnan, Stoica, and Goldberg}]{fox2017multi}
Fox, R.; Krishnan, S.; Stoica, I.; and Goldberg, K. 2017.
\newblock Multi-level discovery of deep options.
\newblock \emph{arXiv preprint arXiv:1703.08294}.

\bibitem[{Heess et~al.(2015)Heess, Wayne, Silver, Lillicrap, Tassa, and
  Erez}]{heess2015learning}
Heess, N.; Wayne, G.; Silver, D.; Lillicrap, T.; Tassa, Y.; and Erez, T. 2015.
\newblock Learning Continuous Control Policies by Stochastic Value Gradients.
\newblock arXiv:1510.09142.

\bibitem[{Konidaris and Barto(2007)}]{10.5555/1625275.1625420}
Konidaris, G.; and Barto, A. 2007.
\newblock Building Portable Options: Skill Transfer in Reinforcement Learning.
\newblock In \emph{Proceedings of the 20th International Joint Conference on
  Artifical Intelligence}, IJCAI'07, 895–900. San Francisco, CA, USA: Morgan
  Kaufmann Publishers Inc.

\bibitem[{Lakshminarayanan et~al.(2016)Lakshminarayanan, Krishnamurthy, Kumar,
  and Ravindran}]{lakshminarayanan2016option}
Lakshminarayanan, A.~S.; Krishnamurthy, R.; Kumar, P.; and Ravindran, B. 2016.
\newblock Option discovery in hierarchical reinforcement learning using
  spatio-temporal clustering.
\newblock \emph{arXiv preprint arXiv:1605.05359}.

\bibitem[{Li(2020)}]{machin}
Li, M. 2020.
\newblock Machin.
\newblock \url{https://github.com/iffiX/machin}.

\bibitem[{Loshchilov and Hutter(2017)}]{loshchilov2017decoupled}
Loshchilov, I.; and Hutter, F. 2017.
\newblock Decoupled weight decay regularization.
\newblock \emph{arXiv preprint arXiv:1711.05101}.

\bibitem[{Machado, Bellemare, and Bowling(2017)}]{10.5555/3305890.3305918}
Machado, M.~C.; Bellemare, M.~G.; and Bowling, M. 2017.
\newblock A Laplacian Framework for Option Discovery in Reinforcement Learning.
\newblock In \emph{Proceedings of the 34th International Conference on Machine
  Learning - Volume 70}, ICML'17, 2295–2304. JMLR.org.

\bibitem[{Mannor et~al.(2004)Mannor, Menache, Hoze, and Klein}]{Mannor04}
Mannor, S.; Menache, I.; Hoze, A.; and Klein, U. 2004.
\newblock Dynamic abstraction in reinforcement learning via clustering.
\newblock \emph{Proceedings of the International Conference on Machine
  Learning}, 21: 560--567.

\bibitem[{McGovern and Barto(2001)}]{McGovern01}
McGovern, A.; and Barto, A. 2001.
\newblock Automatic {D}iscovery of {S}ubgoals in {R}einforcement {L}earning
  using {D}iverse {D}ensity.
\newblock \emph{Proceedings of the International Conference on Machine
  Learning}, 18: 361--368.

\bibitem[{Menache, Mannor, and Shimkin(2002)}]{Menache02}
Menache, I.; Mannor, S.; and Shimkin, N. 2002.
\newblock Q-{C}ut -- {D}ynamic {D}iscovery of {S}ub-{G}oals in {R}einforcement
  {L}earning.
\newblock \emph{Proceedings of the European Conference on Machine Learning},
  13: 295--306.

\bibitem[{Mnih et~al.(2013)Mnih, Kavukcuoglu, Silver, Graves, Antonoglou,
  Wierstra, and Riedmiller}]{mnih2013playing}
Mnih, V.; Kavukcuoglu, K.; Silver, D.; Graves, A.; Antonoglou, I.; Wierstra,
  D.; and Riedmiller, M. 2013.
\newblock Playing Atari with Deep Reinforcement Learning.
\newblock arXiv:1312.5602.

\bibitem[{Nachum et~al.(2018)Nachum, Gu, Lee, and Levine}]{nachum2018near}
Nachum, O.; Gu, S.; Lee, H.; and Levine, S. 2018.
\newblock Near-optimal representation learning for hierarchical reinforcement
  learning.
\newblock \emph{arXiv preprint arXiv:1810.01257}.

\bibitem[{Oh et~al.(2015)Oh, Guo, Lee, Lewis, and Singh}]{oh2015action}
Oh, J.; Guo, X.; Lee, H.; Lewis, R.; and Singh, S. 2015.
\newblock Action-conditional video prediction using deep networks in atari
  games.
\newblock \emph{arXiv preprint arXiv:1507.08750}.

\bibitem[{Oh et~al.(2018)Oh, Guo, Singh, and Lee}]{oh2018self}
Oh, J.; Guo, Y.; Singh, S.; and Lee, H. 2018.
\newblock Self-imitation learning.
\newblock In \emph{International Conference on Machine Learning}, 3878--3887.
  PMLR.

\bibitem[{Pickett and Barto(2002)}]{Pickett02}
Pickett, M.; and Barto, A. 2002.
\newblock PolicyBlocks: {A}n {A}lgorithm for {C}reating {U}seful
  {M}acro-{A}ctions in {R}einforcement {L}earning.
\newblock \emph{Proceedings of the International Conference on Machine
  Learning}, 19: 506--513.

\bibitem[{Puterman(2014)}]{puterman2014markov}
Puterman, M.~L. 2014.
\newblock \emph{Markov decision processes: discrete stochastic dynamic
  programming}.
\newblock John Wiley \& Sons.

\bibitem[{Schaul et~al.(2015)Schaul, Horgan, Gregor, and
  Silver}]{schaul2015universal}
Schaul, T.; Horgan, D.; Gregor, K.; and Silver, D. 2015.
\newblock Universal value function approximators.
\newblock In \emph{International Conference on Machine Learning}, 1312--1320.

\bibitem[{Schaul et~al.(2016)Schaul, Quan, Antonoglou, and
  Silver}]{schaul2016prioritized}
Schaul, T.; Quan, J.; Antonoglou, I.; and Silver, D. 2016.
\newblock Prioritized Experience Replay.
\newblock arXiv:1511.05952.

\bibitem[{Shang et~al.(2019)Shang, Trott, Sheng, Xiong, and
  Socher}]{shang2019learning}
Shang, W.; Trott, A.; Sheng, S.; Xiong, C.; and Socher, R. 2019.
\newblock {Learning World Graphs to Accelerate Hierarchical Reinforcement
  Learning}.
\newblock \emph{arXiv preprint arXiv:1907.00664}.

\bibitem[{Solway et~al.(2014)Solway, Diuk, Cordova, Yee, Barto, Niv, and
  Botvinick}]{Solway14}
Solway, A.; Diuk, C.; Cordova, N.; Yee, D.; Barto, A.~G.; Niv, Y.; and
  Botvinick, M.~M. 2014.
\newblock Optimal behavior hierarchy.
\newblock \emph{PLOS Comp. Bio.}, 10(8).

\bibitem[{Sutton et~al.(2017)Sutton, Modayil, Degris, Pilarski, and
  White}]{sutton2017horde}
Sutton, R.~S.; Modayil, J.; Degris, M. D.~T.; Pilarski, P.~M.; and White, A.
  2017.
\newblock {Horde: A scalable real-time architecture for learning knowledge from
  unsupervised sensorimotor interaction}.

\bibitem[{Sutton, Precup, and Singh(1999)}]{sutton1999between}
Sutton, R.~S.; Precup, D.; and Singh, S. 1999.
\newblock {Between MDPs and semi-MDPs: A framework for temporal abstraction in
  reinforcement learning}.
\newblock \emph{Artificial intelligence}, 112(1-2): 181--211.

\bibitem[{van Hasselt, Guez, and Silver(2015)}]{vanhasselt2015deep}
van Hasselt, H.; Guez, A.; and Silver, D. 2015.
\newblock Deep Reinforcement Learning with Double Q-learning.
\newblock arXiv:1509.06461.

\bibitem[{Veeriah et~al.(2021)Veeriah, Zahavy, Hessel, Xu, Oh, Kemaev, van
  Hasselt, Silver, and Singh}]{abs-2102-06741}
Veeriah, V.; Zahavy, T.; Hessel, M.; Xu, Z.; Oh, J.; Kemaev, I.; van Hasselt,
  H.; Silver, D.; and Singh, S. 2021.
\newblock {Discovery of Options via Meta-Learned Subgoals}.
\newblock \emph{CoRR}, abs/2102.06741.

\bibitem[{Wen et~al.(2020)Wen, Precup, Ibrahimi, Barreto, Van~Roy, and
  Singh}]{NEURIPS2020_4a5cfa92}
Wen, Z.; Precup, D.; Ibrahimi, M.; Barreto, A.; Van~Roy, B.; and Singh, S.
  2020.
\newblock On Efficiency in Hierarchical Reinforcement Learning.
\newblock In Larochelle, H.; Ranzato, M.; Hadsell, R.; Balcan, M.~F.; and Lin,
  H., eds., \emph{Advances in Neural Information Processing Systems},
  volume~33, 6708--6718. Curran Associates, Inc.

\end{thebibliography}

%
%
%

\section{Supplementary Material}
\appendix
\subsection{Additional Empirical Evaluation}
In this section we present additional empirical evaluation of our approach to learn a compression function, complementing the analysis reported in the main text.

Concretely, we evaluate our approach in the MountainCar environment, in which the state consists of the current location and velocity of the agent. In this environment, we collected a replay memory consisting of 200 trajectories of length 200 using a sub-optimal policy that can reach the goal state and can approximately cover all the state space, and learned a compression function with 20 abstract states.

\begin{figure}[h]
    \centering
    \includegraphics[width=8cm]{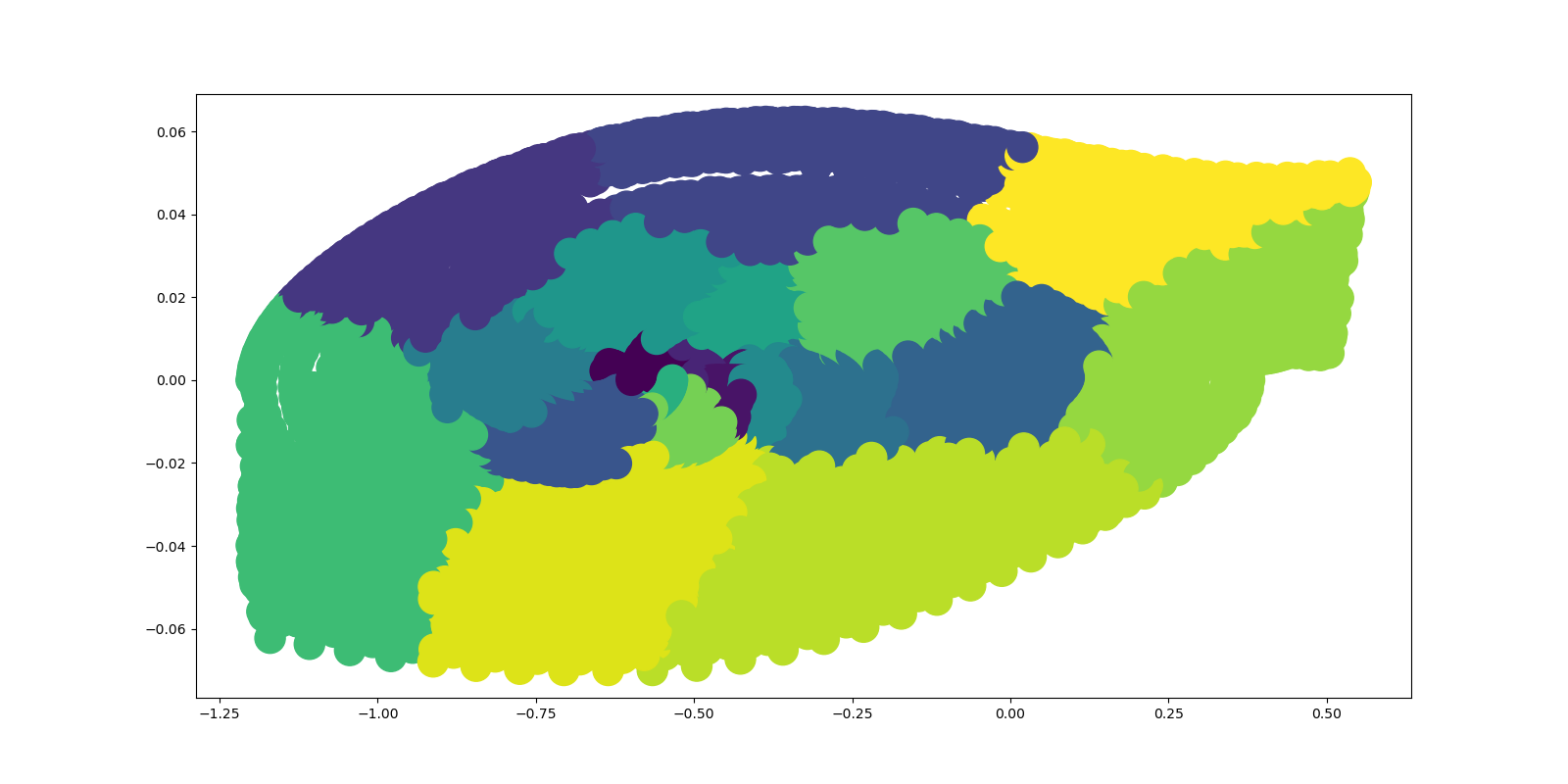}
    \caption{Results of the compression function in the MountainCar environment (axes represent location and velocity); different colors represent different abstract states $z\in Z$}%
    \label{fig:results MountainCar}%
\end{figure}

In ~Figure~\ref{fig:results MountainCar} we show the result of this compression function where different colors represent different abstract states $z\in Z$. Note that the compression function is able to cluster together states that are close in the environment, i.e. states where the car is at similar position and velocity. In particular, states with low velocity near the center are {\em not} very similar to states with high velocity in the same location, and this is captured by the compression function.

\section{Hyperparameters}

Table \ref{fig:HyperparametersHRL} reports the values of the hyperparameters used to train the compression function and the HRL agent.

Table \ref{fig:HyperparametersComp} reports the value of the hyperparameters used to train the DQN-PER and SIL agents.

\begin{table}[H]
\begin{tabular}{|ll|}
\hline
\rowcolor[HTML]{FFFFFF} 
\multicolumn{1}{|l|}{\cellcolor[HTML]{FFFFFF}\textit{\textbf{Hyperparameters}}}                                      & \textit{\textbf{Value}}                                                                                                                                                                                              \\ \hline
\rowcolor[HTML]{FFFFFF} 
\multicolumn{2}{|l|}{\cellcolor[HTML]{FFFFFF}\textit{\textbf{Worker Hyperparameters}}}                                                                                                                                                                                                                                                      \\ \hline
\rowcolor[HTML]{FFFFFF} 
\multicolumn{1}{|l|}{\cellcolor[HTML]{FFFFFF}\begin{tabular}[c]{@{}l@{}}Neural Network \\ Architecture\end{tabular}} & \begin{tabular}[c]{@{}l@{}}CONV1(32, (7, 7), (1, 1))\\ FC1(32)\\ FC2(32)\end{tabular}                                                                                                                                \\ \hline
\multicolumn{1}{|l|}{Activation Function}                                                                            & Relu                                                                                                                                                                                                                 \\ \hline
\rowcolor[HTML]{FFFFFF} 
\multicolumn{1}{|l|}{\cellcolor[HTML]{FFFFFF}Learning rate}                                                          & 0.001                                                                                                                                                                                                                \\ \hline
\multicolumn{1}{|l|}{Optimizer}                                                                                      & Adam                                                                                                                                                                                                                 \\ \hline
\rowcolor[HTML]{FFFFFF} 
\multicolumn{1}{|l|}{\cellcolor[HTML]{FFFFFF}E-Greedy decay}                                                         & 0.9998                                                                                                                                                                                                               \\ \hline
\rowcolor[HTML]{FFFFFF} 
\multicolumn{1}{|l|}{\cellcolor[HTML]{FFFFFF}Batch size}                                                             & 100                                                                                                                                                                                                                  \\ \hline
\multicolumn{1}{|l|}{\begin{tabular}[c]{@{}l@{}}Target network \\ poliak update\end{tabular}}                        & 0.05                                                                                                                                                                                                                 \\ \hline
\multicolumn{1}{|l|}{Discount Factor}                                                                                & 0.95                                                                                                                                                                                                                 \\ \hline
\rowcolor[HTML]{FFFFFF} 
\multicolumn{1}{|l|}{\cellcolor[HTML]{FFFFFF}Replay buffer size}                                                     & $5*10^5$                                                                                                                                                                                                             \\ \hline
\rowcolor[HTML]{FFFFFF} 
\multicolumn{1}{|l|}{\cellcolor[HTML]{FFFFFF}Replay type:}                                                           & PrioritizedExperience Replay                                                                                                                                                                                         \\ \hline
\multicolumn{1}{|l|}{Exponent for prioritization}                                                                    & 0.6                                                                                                                                                                                                                  \\ \hline
\multicolumn{1}{|l|}{Bias Correction}                                                                                & 0.1                                                                                                                                                                                                                  \\ \hline
\rowcolor[HTML]{FFFFFF} 
\multicolumn{1}{|l|}{\cellcolor[HTML]{FFFFFF}\textit{\textbf{Manager Hyperparameters}}}                              &                                                                                                                                                                                                                      \\ \hline
\rowcolor[HTML]{FFFFFF} 
\multicolumn{1}{|l|}{\cellcolor[HTML]{FFFFFF}E-Exploration to learn the model}                                       & 0.995                                                                                                                                                                                                                \\ \hline
\rowcolor[HTML]{FFFFFF} 
\multicolumn{1}{|l|}{\cellcolor[HTML]{FFFFFF}Discount Factor}                                                        & 0.95                                                                                                                                                                                                                 \\ \hline
\rowcolor[HTML]{FFFFFF} 
\multicolumn{2}{|l|}{\cellcolor[HTML]{FFFFFF}\textit{\textbf{Compression Function Hyperparameters}}}                                                                                                                                                                                                                                        \\ \hline
\rowcolor[HTML]{FFFFFF} 
\multicolumn{1}{|l|}{\cellcolor[HTML]{FFFFFF}\begin{tabular}[c]{@{}l@{}}Neural Network \\ Architecture\end{tabular}} & \begin{tabular}[c]{@{}l@{}}CONV1(16, (1, 1), (1, 1))\\ BatchNorm2D(16)\\ CONV2(32, (5, 5), (1, 1))\\ BatchNorm2D(32)\\ CONV3(32, (3, 3), (1, 1))\\ BatchNorm2D(32)\\ FC1(64)\\ BatchNorm1D(64)\\ FC1(1)\end{tabular} \\ \hline
\multicolumn{1}{|l|}{Activation Function}                                                                            & Selu                                                                                                                                                                                                                 \\ \hline
\multicolumn{1}{|l|}{$w_H$, $w_D$ on GridWorld}                                                                      & 0.2, 0.1                                                                                                                                                                                                             \\ \hline
\multicolumn{1}{|l|}{$w_H$, $w_D$ on Mountain Car}                                                                   & 2, 0.1                                                                                                                                                                                                               \\ \hline
\rowcolor[HTML]{FFFFFF} 
\multicolumn{1}{|l|}{\cellcolor[HTML]{FFFFFF}Learning rate}                                                          & 0.001                                                                                                                                                                                                                \\ \hline
\multicolumn{1}{|l|}{Optimizer}                                                                                      & AdamW                                                                                                                                                                                                                \\ \hline
\rowcolor[HTML]{FFFFFF} 
\multicolumn{1}{|l|}{\cellcolor[HTML]{FFFFFF}Batch size}                                                             & \cellcolor[HTML]{FFFFFF}32, 64                                                                                                                                                                                       \\ \hline
\rowcolor[HTML]{FFFFFF} 
\multicolumn{1}{|l|}{\cellcolor[HTML]{FFFFFF}Epochs}                                                                 & 4000                                                                                                                                                                                                                 \\ \hline
\end{tabular}
\caption{Hyperparameters used to train HRL and HRL-Transfer agents}%
\label{fig:HyperparametersHRL}%
\end{table}

\begin{table}[h]
\begin{tabular}{|ll|}
\hline
\rowcolor[HTML]{FFFFFF} 
\multicolumn{1}{|l|}{\cellcolor[HTML]{FFFFFF}\textit{\textbf{Hyperparameters}}}                                      & \textit{\textbf{Value}}                                                                               \\ \hline
\rowcolor[HTML]{FFFFFF} 
\multicolumn{2}{|l|}{\cellcolor[HTML]{FFFFFF}\textit{\textbf{DQN-PER Hyperparameters}}}                                                                                                                                      \\ \hline
\rowcolor[HTML]{FFFFFF} 
\multicolumn{2}{|l|}{\cellcolor[HTML]{FFFFFF}Same as Worker}
\\ \hline
\rowcolor[HTML]{FFFFFF} 
\multicolumn{2}{|l|}{\cellcolor[HTML]{FFFFFF}\textit{\textbf{SIL Hyperparameters}}}                                                                                                                                          \\ \hline
\multicolumn{1}{|l|}{\cellcolor[HTML]{FFFFFF}\begin{tabular}[c]{@{}l@{}}Neural Network \\ Architecture\end{tabular}} & Same as Worker                                                                                        \\ \hline
\rowcolor[HTML]{FFFFFF} 
\multicolumn{1}{|l|}{\cellcolor[HTML]{FFFFFF}Discount Factor}                                                        & \cellcolor[HTML]{FFFFFF}0.95                                                                          \\ \hline
\rowcolor[HTML]{FFFFFF} 
\multicolumn{1}{|l|}{\cellcolor[HTML]{FFFFFF}Replay type:}                                                           & Prioritized Experience Replay                                                                          \\ \hline
\rowcolor[HTML]{FFFFFF} 
\multicolumn{1}{|l|}{\cellcolor[HTML]{FFFFFF}Exponent for prioritization}                                            & 0.6                                                                                                   \\ \hline
\rowcolor[HTML]{FFFFFF} 
\multicolumn{1}{|l|}{\cellcolor[HTML]{FFFFFF}Bias Correction}                                                        & 0.1                                                                                                   \\ \hline
\multicolumn{1}{|l|}{Additional Hyperparameters}                                                                     & \begin{tabular}[c]{@{}l@{}}Same as reported in \\ Self Imitation Learning paper\end{tabular} \\ \hline
\end{tabular}
\caption{Hyperparameters used to train SIL and DQN-PER agents}%
\label{fig:HyperparametersComp}%
\end{table}

\end{document}